# RELATIONSHIP OF THE LANGUAGE DISTANCE TO ENGLISH ABILITY OF A COUNTRY *

**Cao Xinxin**
School of Foreign Studies
Northwestern Polytechnical University,
Xi'an
caoxinxin@mail.nwpu.edu.cn

**Lei Xiaolan**
School of Foreign Studies
Northwestern Polytechnical University,
Xi'an
3031742095@qq.com

**Murtadha Ahmed**
School of Computer Science
Northwestern Polytechnical University
Xi'an
a.murtadha@mail.nwpu.edu.cn

## ABSTRACT

Language difference is one of the factors that hinder the acquisition of second language skills. In this article, we introduce a novel solution that leverages the strength of deep neural networks to measure the semantic dissimilarity between languages based on their word distributions in the embedding space of the multilingual pre-trained language model (e.g., BERT). Then, we empirically examine the effectiveness of the proposed semantic language distance (SLD) in explaining the consistent variation in English ability of countries, which is proxied by their performance in the Internet-Based Test of English as Foreign Language (TOEFL iBT). The experimental results show that the language distance demonstrates negative influence on a country's average English ability. Interestingly, the effect is more significant on speaking and writing subskills, which pertain to the productive aspects of language learning. Besides, we provide specific recommendations for future research directions.

## 1 Introduction

As a special type of human capital, language plays an indispensable role in the economic activities. From the aggregated level, the divergence between languages exerts huge influence on the international bilateral trade (Ku Zussman 2010; Lohmann 2011; Su 2020), migration flows (Adsera Pytlikova 2015), and cross-border market integration (Choi Bordia, 2020; Fenske Kala, 2021). For individuals, making decision of investing in a new language is likely influenced by the expected benefits and the cost associated with the target language (Chiswick Miller 1995). On one hand, multilingual ability facilitates the socioeconomic integration of immigrants in the destination country, and improves their economic status to a large scale (Evans et al. 2020; Zorlu Hartog 2018). On the other hand, the linguistic diversity leads to considerable cost.

In the past decades, language ability has been measured through self-report language proficiency or performance in language proficiency test (Van der Slik, 2010). For instance, many researchers have adopted various methods, e.g., multidimensional scaling and factor analysis, to explain the consistent differences in English among language groups through their TOEFL performance (Elder 1996; Hale, Rock Jirele 1982; Oltman, Stricker Barrows 1988; Swinton Powers 1980). It is Snow (1998) and Kim and Lee (2010) who initiated to empirically evaluate the effects of linguistic and non-linguistic factors on the paper-based and computer-based TOEFL test, TOEFL PBT and TOEFL CBT performance, respectively, for cross-country level. However, the language distance measures were rather limited then, they applied purely qualitative methods to capture the distance between languages. This paper presents a highly quantitative BERT-based semantic language distance (SLD) approach to explain the differentiation in English ability of countries through their average TOEFL iBT score. To the best of our knowledge, we are the first to study the influence of the language distance on a country's English ability through the TOEFL iBT score.

The language distance has been widely discussed in the field of Second Language Acquisition (SLA) (Ellis 1989; Gass, Behney Plonsky 2020; Dekydtspotter, Sprouse Thyre 2000; Diaubalick Guijarro-Fuentes 2019), however, due to the complexity of language, identifying the effects of the language distance appears rather difficult. Therefore, a limited number of empirical strategies, if not non-exist, could facilitate the present analysis on region-based language proficiency. Besides, the existing methods unfortunately cannot capture the semantic similarity between languages, and thus raises the need for an effective method to measure the semantic distance between languages.

With the rapid development of Deep Neural Network (DNN), the Pre-trained Language Model (PLM) (e.g., Word2Vec, Glove, and BERT) based approaches have achieved the state-of-the-art performance of various Natural Language Processing (NLP) tasks, including text generation, machine translation, text classification and sentiment analysis (Qiu et al. 2020; Campos et al. 2019; Peters, Ruder Smith 2019). The ultimate goal of PLM is to semantically map the words that occur in the same context to close points in the latent space. To better illustrate how the words are semantically distributed, we visualize ten words of the closest and the furthest languages (i.e., German and Vietnamese) based on the BERT model in Figure 1. As can be clearly seen, the German words are indeed very close to English than the Vietnamese ones (e.g., 10: kill). In other words, the words that are semantically related are located very close in the embedding space. As these models are well pre-trained in the unsupervised manner (i.e., no labeled data is needed), the embeddings of a pair of languages are readily available. Motivated by this intuition, this study leverages the semantic distance between words of two given languages to measure their language distance. Specifically, it represents each word with a multi-dimensional vector. Then, it uses the cosine similarity to compute the distance of two given wordsfrom different languages.

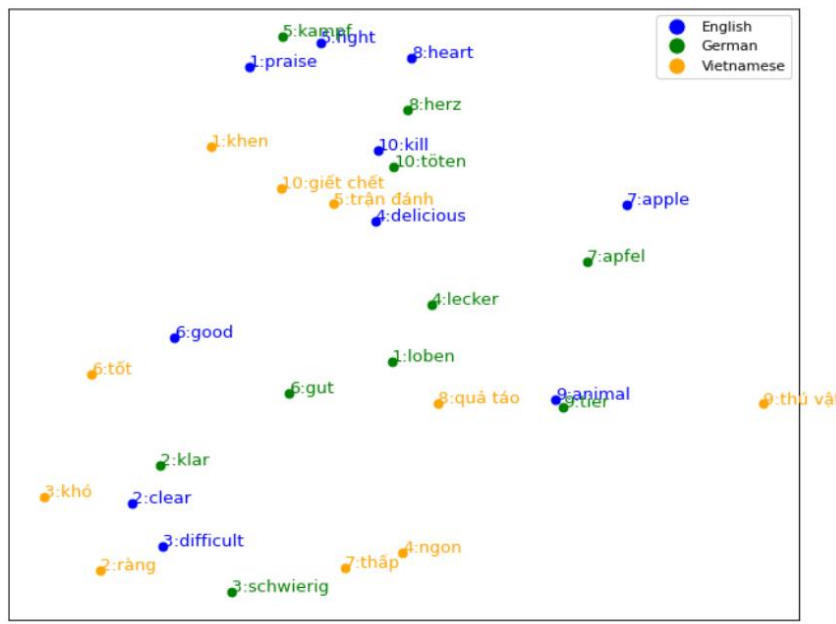

Figure 1. An illustrative example of BERT-based SLD for the words in English, German, and Vietnamese colored with blue, green, and orange, respectively. Each word is labeled with a number, which denotes its respective translation to English.

The main contributions of this paper are summarized as follows:

- A neural network-based approach is introduced to measure semantic dissimilarity between a pair of languages based on their word distributions in the latent space of the pre-trained language model, e.g., BERT;

- We validate the proposed method on the real-data of TOEFL iBT total and subsection scores since the language distance is an important factor impeding the acquisition of new language skills.

The remainder of this paper is organized as follows. Section 2 presents the traditional language distance methods. Section 3 describes the proposed BERT SLD. Section 4 provides the data and the validation process of BERT-based

SLD. Then we provide specific future research directions and conclusions with Section 6.

## 2    Traditional Language Distance Methods

### 2.1    Task Definition

Generally, the language distance denotes the degree of similarity between languages. It is meaningful to distinguish the linguistic distance and the language distance since there is no clear definition of the two terms in literature. The linguistic distance is an umbrella term, which includes the language distance and the perceived linguistic distance (Leusink 2017). The language distance is the objective measure, which can be obtained by comparing features in different languages such as vocabulary, syntax, semantics, phonetic form or grammatical structures. While the subjective side is labeled as the perceived linguistic distance, indicating the distance that learners perceive to exist between languages, but may not actually exist (Kellerman 1983, 1995). Therefore, the major concern of this study comes to the language distance. It should be noted that we only attempt to examine and expand current quantitative measure concerning language distance rather than use the concept of the language distance to operationalize cross-linguistic differences, which largely oversimplifies the current understanding of the multi-dimensional nature of cross-linguistic differences.

### 2.2    Traditional Language Distance Methods

During past several decades, typological linguists and experts in historical and comparative linguistics started to construct quantitative measures to facilitate the analysis of L1 influence on later language learning (Bouckaert, Lemey, Dunn Greenhill 2012; Van der Slik 2010; Campos, Gamallo Alegria 2019). Here, we present two prevailing language distance measures in Linguistics filed, namely, Tree method and ASJP method. Note that there are a bunch of other language distance measures in literature (Van der Slik 2010; Gamallo, Pichel Alegria 2017 etc.), e.g., genetic linguistic distance (Cavalli-Sforza et al. 1994), cognate linguistic distance (McMahon McMahon 2005), score-based measure (Chiswick and Miller 1998; Hart-Gonzalez and Lindemann 1993), World Atlas of Language Structures (WALS) measure ((Lohmann 2011) etc., which are excluded in this study either because of their widely-proved disputability or the limited coverage of solely Indo-European languages (Isphording Otten 2014; Van der Slik 2010; etc.).

### 2.2.1    Tree Method

Apart from the WALS index, there exists another source of language information called the Ethnologue[1] project, which aims at evaluating the family relations for all existing languages (Campbell 2008). Tree approach was a purely qualitative method developed in phylogenetics, a subfield of historical and comparative linguists, to retrace the evolution of the world languages. However, it has become an alternative measure for the language distance since the concept of language proximity index was introduced by Adsera and Pytlikova (2015). For language proximity, 1 denotes the same language and 0 stands for languages without any family relations. In the middle of the two values, it assigns a value, e.g., 0.1, 0.25, 0.45, and 0.7, denoting the number of shared branches. Then, the language distance is defined by LD = 1 value. Regardless of the expert opinions behind, tree-based language distance value shows only a marginal variation. What's more, Ethnologue project is not freely available now.

### 2.2.2    ASJP Method

The Automatic Similarity Judgment Program (ASJP[2]) method has achieved a considerable improvement in performance compared to the aforementioned alternatives. It was implemented based on the findings of German Max Planck Institute for Evolutionary Anthropology. ASJP was originally designed for generating language trees to automatically compute language distance between languages based on their phonetic representations. Specifically, it begins with a list of 40 words, e.g., parts of human body, selected from the Swadesh corpus (Swadesh 1952) and then transfers them into their phonetic script, called ASJP code. For instance, the person pronoun I is transferred into Ei in English, while wataSi in Japanese. Then, it explores Levenshtein Distance, i.e., the minimum edit distance between two strings, to calculate both the local (each word pair) and global (non-related items) distance existed between two phonetic strings. The resulting normalized and divided Levenshtein distance (LDND) is interpreted as the degree of phonetic dissimilarity between two languages (see details in Isphording Otten 2014). Although ASJP method has been proved internationally to be a powerful method for computing language distance, it captures only the phonetic differences between languages.

Overall, the aforementioned measures for the language distance are very helpful for comparing relatively peripheral language pairs with restricted number of observations. However, they also demonstrate some disadvantages in their own calculating processes, e.g., across the two measures, the language distance is derived on the small data comparison.

---

[1] Ethnologue: https://www.ethnologue.com/.

[2] ASJP: https://asjp.clld.org/.

Besides, these methods unfortunately cannot capture the semantic distance between languages, which is of great value in Linguistic studies (Robins 2014; Palmer Frank Robert 1981).

## 3 BERT SLD Method

As BERT is curial component of our approach, this section highlights the basic concepts.

### 3.1 Bidirectional Encoder Representations from Transformers (BERT)

Bidirectional Encoder Representations from Transformers (BERT) (Devlin et al. 2018) is the state-of-the-art language representation model. BERT is built upon the Transformer architecture and released in the tensor2tensor library (Vincent et al. 2008; Devlin et al. 2018). Two basic steps are involved: pre-training and fine-tuning. BERT was pre-trained on unlabeled large corpora from Wikipedia and Google book. In order to train a deep bidirectional representation, two tasks were performed. Masked Language Model (MLM) task, which randomly masks a percentage of the input tokens, and then requires the model to predict them, and Next Sentence Prediction (NSP) task that encourages BERT to learn the relationship between sentences. For further fine-tuning step, BERT has been used widely in various downstream tasks of NLP (e.g., sentiment analysis, topic classification, etc.) (Karimi, Rossi Prati 2020; Devlin et al. 2018). Specifically, they initialize the weights of the network by the pre-trained weights of BERT and then rely on the labeled corpus to fine-tune these weights in order to improve the performance on the task on-target. Despite the significant improvement of these methods, the performance heavily relies on the labeled data, which may not readily available in real-world scenarios.

In our work, we simply exploit BERT at an inference step to benefit from the success of contextual representation (i.e., word distribution) to compute the semantic distance between two texts of different languages. In other words, we rely on the pre-trained weights of BERT without further fine-tuned step for two reasons: (1) Our approach simply computes the semantic distance between two texts based on their corresponding weights generated by BERT; (2) Fine-tuning step requires an additional manually annotated bilingual corpus and a defined downstream task (i.e., natural language inference), which is labor-intensive and may not readily available in real-world scenarios. In our experiment, we use mBERT[3] (a new version of BERT for multilingual learning) as a weight generator. Specifically, the output of the language model is a matrix $E \in R^{n \times d}$, where $n$ is the number of words in the vocabulary and $d$ is the dimension of embedding, e.g., 768. Then, we retrieve the corresponding weights of a bilingual corpus Bible to compute the semantic distance.

### 3.2 Bible Corpus

In our implementation, we use a multilingual parallel corpus created from 100 translations of the Bible[4]. Several characteristics are worth mentioning about the Bible corpus. First, with the globally missionary expansion, the Bible has more than 100 translated versions, which exceeds any other single work in literature (Christodouloupoulos Steedman 2015). The highly parallel corpus offers excellent and valuable information for low-resource language pairs. Next outstanding feature of this multilingual corpus comes to the text size of the Bible. The average size of normal fiction novels is about 100k words, while the Old Testament and the New Testament together contain around 800k English words (Christodouloupoulos Steedman 2015), enabling adequate feed to the pre-training of the model. Besides, the structure of the Bible is also unique as it is divided into books, chapters and verses, which allows for the model to automatically align every language at the verse level without ambiguity. Another characteristic we should note is that the translating principle behind the Bible is basically "sense-for-sense" since the aim of most missionary linguists was to convey the message from God as accurate as possible. Thus, the semantic meaning of the Bible is well kept in almost every language with this content-sensitive translation approach (Christodouloupoulos Steedman 2015).

### 3.3 BERT SLD computation

We leverage the PLM BERT, which has achieved the state-of-the-art performance in various NLP tasks (Gamallo et al. 2017). The key idea behind it is to map the words that occur in the same context into close points in the latent space, in other words, to convert the input words into the semantic vectors. With the bilingual dictionaries provided on Github community https://anonymous.4open.science/r/bible-corpus-1FF2, now we describe how to use the resulted vectors to measure the semantic language distance.

---

[3] mBERT: https://anonymous.4open.science/r/bert-DA06.

[4] Bible-corpus: https://anonymous.4open.science/r/bible-corpus-1FF2.

### 3.3.1 Cosine Similarity

Given bilingual dictionary, the output of the language model is a matrix $E \in R^{n \times d}$, where $n$ is the number of words in the vocabulary and $d$ is the dimension of embedding, e.g., 300. Thus, each row is a $d$ vector that represents the respective word distribution. Given two words of different languages, we compute the cosine similarity between their respective vectors as follows:

$$score(w_i, w_j) = f(e_i, e_j) \tag{1}$$

where $e_I$ and $e_J$ denote the respective vectors of source word $w_I$ and the target word $w_J$, respectively, and $f(.)$ represents the cosine similarity $\vartheta$ function. Then, we average the scores of the bilingual dictionary of a given pair of language $k$ to compute the semantic similarity between languages:

$$SLS(l_k) = \frac{1}{n} \sum score(w_i, w_j), \tag{2}$$

in which SLS(.) can be interpreted as how words of given language $l_k$ are close to their parallel English ones in the latent space. As the ultimate goal is to compute the semantic distance, we coast the semantic similarity to the semantic distance as follows:

$$SLD = 1 - SLS \tag{3}$$

The result of the language distance derived from both PLM BERT and the traditional methods are presented in Table 1.

Table 1. Language Distance between 33 Languages and English

| Language | Language Family | BERT | ASJP | Tree |
|----------|----------------|------|------|------|
| Albanian | Indo-European | 0.16 | 0.95 | 0.90 |
| Arabic | Afro Asiatic | 0.23 | 0.99 | 1.00 |
| Armenian | Indo-European | 0.36 | 0.97 | 0.90 |
| Bulgarian | Indo-European | 0.16 | 0.87 | 0.90 |
| Burmese | Sino Tibetan | 0.33 | | 1.00 |
| Chinese | Sino Tibetan | 0.28 | 1.00 | 1.00 |
| Croatian | Indo-European | 0.20 | 0.87 | 0.90 |
| Czech | Indo-European | 0.18 | 0.91 | 0.90 |
| Danish | Indo-European | 0.18 | 0.67 | 0.90 |
| French | Indo-European | 0.16 | 0.89 | 0.90 |
| German | Indo-European | 0.16 | 0.69 | 0.55 |
| Greek | Indo-European | 0.22 | 0.97 | 0.90 |
| Hebrew | Afro Asiatic | 0.23 | 0.98 | 1.00 |
| Hungarian | Uralic | 0.24 | 0.95 | 1.00 |
| Indonesian | Austronesian | 0.37 | 0.99 | 1.00 |
| Japanese | Japonic | 0.38 | 1.01 | 1.00 |
| Korean | Koreanic | 0.27 | 1.00 | 1.00 |
| Lithuanian | Indo-European | 0.19 | 0.95 | 0.90 |
| Nepali | Indo-European | 0.32 | 0.99 | 0.90 |
| Norwegian | Indo-European | 0.19 | 0.64 | 0.75 |
| Persian | Indo-European | 0.21 | 0.92 | 0.90 |
| Portuguese | Indo-European | 0.15 | 0.90 | 0.90 |
| Romanian | Indo-European | 0.17 | 0.87 | 0.90 |
| Russian | Indo-European | 0.17 | 0.95 | 0.90 |
| Serbian | Indo-European | 0.20 | 0.00 | |
| Slovak | Indo-European | 0.19 | 0.92 | 0.90 |
| Slovenian | Indo-European | 0.17 | 0.91 | 0.90 |
| Spanish | Indo-European | 0.17 | 0.91 | 0.90 |
| Swedish | Indo-European | 0.18 | 0.62 | 0.55 |
| Thai | Tai Kadai | 0.21 | 0.99 | 1.00 |
| Turkish | Turkic | 0.34 | 0.98 | 1.00 |
| Ukrainian | Indo-European | 0.36 | 0.94 | 0.90 |

| Vietnamese | Austroasiatic | 0.22 | 1.04 | 1.00 |

*Notes.* Due to the unavailability of several language pairs, there are some missing values indicated by the space in the above table.

## 4 Validation Evidence

In this section, we aim to examine the predictive validity of the proposed language distance method on the average English ability of a country by investigating the relationship of BERT-based language distance to TOEFL iBT performance. First, we describe the data used in our implementation. Second, we examine the relationship between BERT-based SLD and TOEFL iBT total and subsection scores through two steps of analysis: (1) correlation of BERT-based language distance with TOEFL iBT total and subsection scores, and correlation of ASJP, Tree language distance measures with TOEFL iBT total and subsection scores; and (2) Multivariate Analysis of Variance (MANOVA) model, an extension of ANOVA, is employed to further evaluate the effects of BERT-based language distance on TOEFL iBT total and subsection scores. Both correlational analysis and MANOVA analysis are performed through IBM SPSS Version 23 (IBM Corp 2018).

### 4.1 Data

Test of English as a Foreign Language (TOEFL) is designed to measure the English proficiency of people whose mother language is not English. According to Education Testing Service board (ETS 2021), TOEFL has been accepted for admission purposes by over 11000 institutions, including universities, colleges and even governments in more than 150 countries, which significantly proves its efficacy in measuring the language proficiency in English. TOEFL test underwent two big changes to better assess language learning. The original paper-based TOEFL (TOEFL PBT) was introduced in 1962, which consists of three parts: reading, listening and speaking tasks. With the release of computer-based TOEFL (TOEFL CBT) in 1998, writing becomes one component of TOEFL test. Next big improvement was the introduction of internet-based TOEFL (TOEFL iBT) in September 2005. The TOEFL iBT integrates all the four language skills, namely reading, listening, speaking and writing, which are designed to measure the ability to understand academic written materials, the ability to understand spoken English in the academic setting, the ability to speak in English and the ability to write in English, respectively.

The total score of TOEFL iBT is 120 points, and each of the four subsections occupies 30 points. A sample of TOEFL iBT test paper can be found on TOEFL website, where readers can take a closer look at the structure, administration and detailed requirements of the test. For scoring, reading and listening parts are scored through automatic AI scoring, while speaking and writing sections are done by combination of computer and highly experienced, well-trained human raters to minimize the possible rater bias and provide a more comprehensive and accurate evaluation of one's ability.

Table 2. An example of the TOEFL iBT dataset in 2019: TOEFL iBT total and section score means with all examinees classified by native country

| Native Country | Reading | Listening | Speaking | Writing | Total |
|---|---|---|---|---|---|
| Germany | 24 | 26 | 25 | 24 | 98 |
| Hungary | 23 | 24 | 23 | 22 | 92 |
| Ukraine | 20 | 22 | 22 | 21 | 86 |
| Turkey | 20 | 21 | 20 | 20 | 80 |
| Saudi Arabia | 16 | 20 | 21 | 18 | 74 |

*Note.* Mean scores for small samples of less than 30 test-takers are not reported to ensure the reliability. In addition, due to rounding, section scores may not add up to the total scores.

In each year, TOEFL board releases the Test and Score Data Summary for TOEFL iBT, where total and subsection score means from January to December with all examinees classified by country or region are available (see Table 2: An example of the TOEFL iBT dataset in 2019). The fair and objective scoring process makes this data reliable for comparing performance of examinees from different geographic regions or countries. To guarantee the reliability and validity, TOEFL website does not report means for small groups of less than 30 participants. Although the starting point of ETS board is not to encourage the practice of ranking countries, the surprisingly consistent and recurring variation pattern in scores does shed some light on the overall English proficiency profiles of specific regions or countries. As can be seen in Table 3, across three years, European participants constantly achieve higher scores than participants from other continents, especially Africa. Likewise, in Table 2, it is noticeable that both total and subsection scores of German examinees exceed that of Saudi Arabia to a large extent, which is in accordance with their real English proficiency profiles. In this study, given the impacts of COVID-19 pandemic on 2020 TOEFL iBT test (Ockey 2021), we leverage

TOEFL iBT Test and Score Data Summary from 2017 to 2019 as our data. Around 170 countries attend TOEFL iBT test each year, however, we can investigate on the possible effects of the language distance only on the condition that the target country has one official language, otherwise we need to assign two values to a single sample, which is infeasible. Therefore, excluding those bilingual or multilingual nations, and further considering the BERT-based SLD coverage, our sample is composed of 91 countries where in total 34 different languages (including English) are spoken (more details can be found in the Supporting Online Information). To best showcase the effect of the language distance, the target sample covers languages from 9 language families including Indo-European, Afro-Asiatic, Sino-Tibetan, Uralic, Austronesian, Japonic, Koreanic, Tai Kadai and Turkic. Recent studies have consistently shown that TOEFL total scores do not constitute a meaningful indication of language proficiency. Instead, the subscores are more meaningful (e.g., Ginther Yan 2018; Cho Bridgment 2012). If one only examines the total scores, it is possible that these fine-grained differences will be masked. Therefore, we include both the TOEFL total and subscale scores to enable a more accurate operationalization of English language proficiency.

Table 3. TOEFL iBT total and section score means of continents by year

|  | Reading | Listening | Speaking | Writing | Total |
|---|---|---|---|---|---|
| 2019 |  |  |  |  |  |
| AFRICA | 17 | 19 | 20 | 19 | 74 |
| AMERICAS | 20 | 22 | 22 | 21 | 84 |
| ASIA | 20 | 21 | 21 | 21 | 82 |
| EUROPE | 22 | 24 | 23 | 22 | 91 |
| MIDDLE EAST | 19 | 21 | 22 | 20 | 81 |
| PACIFIC REGION | 22 | 24 | 23 | 22 | 91 |
| 2018 |  |  |  |  |  |
| AFRICA | 16 | 18 | 20 | 19 | 74 |
| AMERICAS | 20 | 22 | 22 | 21 | 84 |
| ASIA | 19 | 21 | 21 | 21 | 82 |
| EUROPE | 21 | 23 | 23 | 22 | 90 |
| MIDDLE EAST | 18 | 21 | 22 | 20 | 80 |
| PACIFIC REGION | 21 | 22 | 22 | 22 | 86 |
| 2017 |  |  |  |  |  |
| AFRICA | 17 | 18 | 20 | 19 | 74 |
| AMERICAS | 20 | 22 | 22 | 21 | 85 |
| ASIA | 19 | 20 | 21 | 21 | 81 |
| EUROPE | 22 | 23 | 23 | 22 | 90 |
| MIDDLE EAST | 18 | 21 | 22 | 20 | 80 |
| PACIFIC REGION | 20 | 21 | 22 | 21 | 83 |

## 4.2 Results and Discussion

### 4.2.1 Descriptive Statistics

Table 4 presents the basic descriptive statistics for TOEFL iBT total and subsection score means of 91 countries by year. Both the average TOEFL iBT total and subsection scores show little variation across three years with the mean clustering around 80 points. However, it is interesting to note that the same score profile pattern recurs in all three years: speaking> listening> writing> reading, which represents the typical TOEFL score profile or English language skills' profile of most countries in our sample. This specific pattern still exists when it comes to the average score of all countries attending that TOEFL iBT test in that year. A multitude of reasons can account for the outscore of the oral sections than the written sections. First, from the perspective of TOEFL administration, reading section is more time-limited than the other parts, examinees are required to read three to four academic passages of more than 700 words and respond to 30-40 multiple choice questions in around 60 minutes (TOEFL 2022). Based on the ETS standard of mapping TOEFL iBT test scores into Common European Framework Reference (CEFR) (TOEFL 2022) (see Table 5), the cut score of speaking section at C1, B2, B1 and A2 levels remains highest among the four subsections, which partly suggests that speaking section might be easier or examinees tend to achieve higher scores is the speaking part. Second, from the aspect of participant characteristics, students from different regions demonstrate different score profiles associated with that place (Ginther Yan 2018). Oral part never appears difficult for those who are frequent users of English, e.g., Americans, while in countries like China, examinees are more competent at written parts of the exam, i.e., reading and writing, as shown in the Test and Score Data Summary each year. Countries with score profile of higher oral competence and lower written competence outnumbers those having conversed score profile, thus leading to the intriguing pattern discussed above.

Table 4. Descriptive statistics for TOEFL iBT total and subsection score means of studied samples by year

|  | 2019 (N=91) | | 2018 (N=91) | | 2017 (N=91) | |
| --- | --- | --- | --- | --- | --- | --- |
|  | M | SD | M | SD | M | SD |
| Total | 84.70 | 6.842 | 84.10 | 7.124 | 84.13 | 7.317 |
| Reading | 20.10 | 2.166 | 19.70 | 2.178 | 20.00 | 2.196 |
| Listening | 21.81 | 2.016 | 21.46 | 2.182 | 21.41 | 2.231 |
| Speaking | 21.98 | 1.626 | 21.85 | 1.763 | 21.79 | 1.877 |
| Writing | 20.87 | 1.600 | 20.89 | 1.609 | 21.02 | 1.725 |

Table 5. Mapping TOEFL iBT scores into CEFR levels

| CEFR level | Total | Reading | Listening | Speaking | Writing |
| --- | --- | --- | --- | --- | --- |
| C2 | 114 | 29 | 28 | 28 | 29 |
| C1 | 95 | 24 | 22 | 25 | 24 |
| B2 | 72 | 18 | 17 | 20 | 17 |
| B1 | 42 | 4 | 9 | 16 | 13 |
| A2 | N/A | N/A | N/A | 10 | 7 |

*Note.* N/A: not applicable or not available.

### 4.2.2 Correlational Analysis

We provide Pearson correlations between language distance and TOEFL iBT total and subsection scores in Table 6. The relationship between the language distance and English proficiency is operationalized by the correlation coefficients between language distance values and TOEFL iBT test scores. Many scholars have repeatedly pointed out, there is no solid guidelines with regard to the interpretation of the correlation coefficients, especially in studies related to language tests because the dependent variable (e.g., language proficiency test scores) is often determined by a number of factors besides the studied independent variable (e.g., the language distance) (e.g., Ginther Yan 2018; Kim Lee 2010; Van der Slik 2010). For any single independent variable, it is impossible to observe a correlation coefficient that indicates a high effect, e.g., 0.75. In such case, the interpretation should be based on the expected magnitude of strength, even a small correlation coefficient, e.g., 0.3, can be perceived as strong and substantially meaningful (e.g., Cho  Bridgment 2012; Ginther Yan 2018; Rosenthal Rubin 1982; Sackett, Borneman Connelly 2008). In our study, as expected, consistent negative correlation between the language distance and TOEFL iBT total and section scores occurs in year 2017, 2018 and 2019, and correlational patterns are similar across the three years.

Table 6. Pearson correlations between language distance and TOEFL iBT scores

|  | Reading | Listening | Speaking | Writing | Total |
| --- | --- | --- | --- | --- | --- |
| **2019** | | | | | |
| BERT | -0.015 | -0.142 | -0.390*** | -0.316** | -0.207* |
| ASJP | -0.011 | -0.118 | -0.381*** | -0.391*** | -0.220* |
| Tree | -0.032 | -0.137 | -0.385*** | -0.408*** | -0.237* |
| **2018** | | | | | |
| BERT | -0.047 | -0.148 | -0.355*** | -0.271** | -0.208* |
| ASJP | -0.053 | -0.130 | -0.340*** | -0.340*** | -0.218* |
| Tree | -0.070 | -0.146 | -0.346*** | -0.354*** | -0.233* |
| **2017** | | | | | |
| BERT | -0.117 | -0.194 | -0.394*** | -0.333*** | -0.274** |
| ASJP | -0.089 | -0.173 | -0.385*** | -0.396*** | -0.269* |
| Tree | -0.107 | -0.187 | -0.392*** | -0.407*** | -0.283** |

*Note.* *p<0.05, **p<0.01, ***p<0.001, two-tailed.

In 2019 cohort, an interesting correlation pattern between the language distance methods and TOEFL iBT scores emerges. As presented in Table 6, the correlation between BERT-based SLD, ASJP, Tree method and TOEFL iBT total score gathers around 0.2 (r total =-0.207*, r total =-0.220*, r total =-0.237*, respectively). However, when cut down into subsection scores, the correlation pattern demonstrates much variability. Strongly negative correlation can be observed in both speaking and writing subsections, and the magnitude is similar across BERT-based SLD, ASJP and Tree method (r speaking=-0.390***, r speaking=-0.381***, r speaking=-0.385***; r writing=-0.316**, r writing=-0.391***, r writing=-0.408***). For reading and listening section, the effect of this negative correlation decreases in a great scale, especially in reading section, where correlation coefficients are close to zero (r reading=-0.015, r reading=-0.011, r reading=-0.032). Similar patterns appear in 2018 and 2017 cohort, but the effect size shows a minor fluctuation. The overall correlation coefficients in the year 2018 are slightly lower than 2019, while those of 2017 cohort are a bit higher.

Overall, across three years, there is a moderately strong negative correlation observed between the language distance and TOEFL iBT total score. Furthermore, consistent and strong negative correlation is observed in case of TOEFL speaking and writing subsections. However, the correlation coefficients drop substantially in listening subscale, and the effect size shows no significance. For reading subscale, the correlation even disappears. TOEFL iBT total and subsection scores demonstrate great variation in terms of correlation magnitude, which suggests that section score might be a more meaningful and accurate measure of language skill than the total test score, as is recently proved by Ginther Yan (2018). The analysis of Pearson correlation coefficients seems to indicate that the language distance can exert more impacts on the productive skills (speaking and writing) than the receptive skills, i.e., reading and listening. However, such interpretation should be supported by careful and thorough investigation.

### 4.2.3 MANOVA Analysis

To determine whether there is a statistically significant difference in TOEFL iBT performance between groups that have varying degrees of closeness to English, we categorize the samples into groups by language distance, e.g., BERT language distance values range from 0 to 0.38, those with 0-0.19 are clustered into Group A, and 0.20-0.38 into Group B, thus languages in Group A are considered as closer to English while Group B shows more distance to English. Similar operation is conducted for ASJP and Tree cohorts, yet there is a slight difference in manipulation. ASJP values range from 0 to 1.04, however, there is no value between 0 and the 0.62, which doubtlessly would cause the imbalance in sample size between two groups, therefore, we further make a cutline between 0.62-1.04, those fall into 0-0.83 are categorized into Group A, and those range from 0.84 to 1.04 in Group B. Operation in Tree cohort is the same of ASJP method. Comparison of the mean scores of TOEFL iBT test between two groups is then performed through one-way MANOVA, a technique used to determine the influence of independent categorical variables on multiple dependent variables, e.g., total score and four subsection scores in the current study. It is slightly different from one-way ANOVA model in that it allows for simultaneous evaluation of more than one predicted variable. Quantile-Quantile (Q-Q) Plot shows that the predicted variables are normally distributed across three years, and equality of variance is guaranteed through the Homogeneity of Variance Test. Post hoc test is not performed because there are fewer than three groups in this study. In subsequent sections, we present and discuss only the 2017 cohort since there is little variation across three years.

Table 7. MANOVA results: the effects of BERT language distance on 2017 TOEFL iBT scores

|            | Reading | Listening | Speaking | Writing | Total |
|------------|---------|-----------|----------|---------|-------|
| $Mean_A$   | 20.56   | 21.96     | 22.31    | 21.41   | 86.19 |
| $Mean_B$   | 19.19   | 20.59     | 21.03    | 20.46   | 81.14 |
| F          | 9.283   | 8.993     | 11.538   | 7.074   | 11.703 |
| **Sig**    | **0.003** | **0.004** | **0.001** | **0.009** | **0.001** |

*Note.* Significance level: 0.05

Table 8. MANOVA results: the effects of ASJP language distance on 2017 TOEFL iBT scores

|            | Reading | Listening | Speaking | Writing | Total |
|------------|---------|-----------|----------|---------|-------|
| Mean_A     | 20.05   | 21.78     | 23.10    | 22.35   | 87.00 |
| Mean_B     | 20.00   | 21.32     | 21.55    | 20.76   | 83.57 |
| F          | 0.011   | 0.557     | 7.924    | 11.382  | 2.751 |
| Sig        | 0.916   | 0.457     | 0.006    | 0.001   | 0.101 |

*Note.* Significance level: 0.05

The MANOVA results of BERT, ASJP and Tree in 2017 cohort are presented in Table 7, Table 8 and Table 9, respectively.

In BERT case (see Table 7), MANOVA results indicate that there is a surprisingly significant variation between the means of Group A and B, across the total score and four subsections (p reading=0.003, p listening=0.004, p speaking=0.003, p writing=0.003p total=0.003, respectively). In terms of ASJP and Tree method, on contrary, as shown in Table 8 and Table 9, for TOEFL total score means, no statistically meaningful difference can be observed. For reading and listening section, the mean scores between Group A and B are quite similar, and statistically significance does not exist. More interestingly, for speaking and writing section, the mean scores differences demonstrate a huge variation, especially in writing section, the effect between Group A and Group B seems mildly stronger than that of BERT case (p asjp=0.003, p tree=0.003). In summary, when two groups are divided by ASJP or Tree measure, meaningful variation can be observed only in speaking and writing sections. However, there is consistent and significant difference between two groups clustered by BERT language distance in terms of both TOEFL total and four subscale means, which partly suggests that BERT language distance is more advantageous and suitable in explaining TOEFL iBT scores. Although the four TOEFL subsections, i.e., reading, listening, speaking and writing, are initially designed to measure four separate subskills in language learning, they are integrated to some extent in that one cannot accomplish any single task in each of the four sections without a full comprehension of the semantic meaning of the context (TOEFL 2022). Pearson correlation coefficients suggest that there are strong positive correlations between BERT based language distance and ASJP, between BERT and Tree method (r=0.878**, r=0.866**, respectively), yet their focuses are quite distinctive. What BERT model attempts to capture is exactly the semantic differences between different languages. Therefore, it is reasonable and justifiable that BERT based language distance demonstrates more robustness in explaining the TOEFL score variation.

The results of MANOVA seem to support the hypothesis that the language distance is a strong predictor of TOEFL iBT productive skills (reflected in speaking and writing). With respect to receptive skills (reflected in reading and listening), the predictive power varies across different language distance computation methods. For most learners, receptive or passive skills are the first step in learning a second language, which usually functions as a springboard to the acquisition of productive or active skills (Rico 2014). In the initial input stage of language learning, the language distance plays a minor role. Yet in the subsequent output stage of language learning, i.e., speaking and writing, the role of the language distance becomes more visible and significant. As illustrated above, across three entirely independent measurement of language distance, i.e., BERT, ASJP and Tree, the significantly negative meaningful correlation remains between the language distance and TOEFL iBT speaking and writing scores, which suggests that language distance is likely to hinder the learning of productive competence. The negative impact of the language distance on the learners' productive skills was also confirmed in one study conducted by Van der slik (2010), he examined the influence of mother tongue on immigrants' language proficiency in Dutch. Immigrants with mother tongues not belonging to Indo-European languages are proved to be less capable in Dutch speaking and written proficiency. However, we need to be rather careful and cautious when interpreting such results.

Table 9. MANOVA results: the effects of Tree language distance on 2017 TOEFL iBT scores

|         | Reading | Listening | Speaking | Writing | Total  |
|---------|---------|-----------|----------|---------|--------|
| Mean_A  | 20.07   | 21.80     | 23.00    | 22.33   | 87.00  |
| Mean_B  | 19.99   | 21.31     | 21.53    | 20.76   | 83.52  |
| F       | 0.016   | 0.605     | 8.197    | 11.488  | 2.859  |
| Sig     | 0.899   | 0.439     | 0.005    | 0.001   | 0.094  |

*Note.* Significance level: 0.05

Language proficiency test itself is complicated because it is determined by multi-layers of factors. For instance, in terms of country-level English proficiency, score profile would be one of the most powerful predictors. Many countries show strengths and weaknesses in different subskills; and depending on the subskill (e.g., speaking vs. writing), there can be complex combinations of subskill profiles (e.g., subscore profiles of students from the Middle East vs. China). When adding up the effects of these profiles, specific patterns, e.g., listening>speaking>reading>listening, are likely to emerge. Additionally, factors such as the average schooling time (Kim 2010; Snow 2008), education quality, the average investment in English teaching and learning (Hakuta 1976), experience of being ex-colony of Britain or U.S.( Migge Léglise 2008; Spencer 2017), countries' reliance on business with English-speaking countries, linguistic diversity in the native country, cultural differences (Muthukrishna, et al. 2020), issue of lingua franca, power of the native language spoken in the country etc., all could be meaningful predictive indicators of a country's English proficiency. In addition, TOEFL administration features, task difficulty of each subtask, preparation efficacy might also partly contribute to the aforementioned phenomenon. Therefore, in future studies, more empirical evidence and stronger theoretical justification are expected to support the interpretation of the relationship between the language distance and the predicted English proficiency.

## 5 Conclusion and Future Work

This study presents a relatively new neural network model BERT-based approach to quantitatively measure the semantic distance between languages, and further validate the method on the real-data of TOEFL iBT scores since the language distance is an important factor hindering individuals in specific geographical regions to acquire new language skills. The experimental results have shown that the current quantified language distance measures can negatively influence the TOEFL iBT speaking and writing subskills, which pertain to the productive aspect of language learning. Additionally, the introduced BERT based language distance method outperforms the existing methods in predicting both the TOEFL iBT reading, listening subsection scores and total score.

A potential drawback of this study is that BERT based language distance computation indeed raises some technical challenges for SLA researchers. It is noteworthy that BERT was trained on large corpora, and thus reflects on the word's distribution in its latent space. The languages that share large parallel corpora may have better mapping of semantic feature, thus influences negatively on the language distance computation (Davison, Feldman Rush 2019; Rama, Beinborn Eger 2020; Peters, Ruder Smith 2019). For future work, a regularization term is highly recommended to discourage the model learn weights based on amount of data, technical details, however, need further investigation.

### Acknowledgments

We would like to thank AI community for providing valuable open-source information about BERT model. We would also like to thank the Educational Testing Service center (ETS) who offers us authoritative and valuable information about the TOEFL iBT test.